\documentclass[runningheads, envcountsame, a4paper]{llncs}
\usepackage{csquotes}
\usepackage{subfig}
\usepackage{mathtools}
\usepackage{amsmath}
\usepackage{amssymb}
\usepackage{multirow}
\usepackage[misc]{ifsym}
\usepackage{standalone}
\usepackage{commath}
\usepackage{tikz}
\usepackage{csquotes}
\usetikzlibrary{shapes.geometric,arrows, fit,positioning}

\usepackage{mathtools}

\usepackage{booktabs}
\usepackage{multirow}
\usepackage{array}
\usepackage{adjustbox}
\usepackage{hhline}

\usetikzlibrary{fit}

\usepackage{float}

\usepackage{tasks}
\usepackage{dblfloatfix}
\usepackage{supertabular}

\usepackage{hhline}

\usepackage{graphicx}
\usepackage[hyphens]{url}

\usepackage{longtable} 
\usepackage{multirow} 

\usepackage{commath,amsmath, amsfonts, amssymb}
\usepackage{mathtools, dsfont}

\usetikzlibrary{shapes.geometric,arrows, fit,positioning}
\usetikzlibrary{arrows, decorations.markings}
\usetikzlibrary{arrows.meta,calc}
\usetikzlibrary{calc}
\usetikzlibrary{decorations.pathreplacing}
\usetikzlibrary{matrix,backgrounds,positioning}
\usetikzlibrary{decorations.pathmorphing}
\usetikzlibrary{shapes.misc}

\usepackage{xcolor}
\usetikzlibrary{shadows.blur}
\usetikzlibrary{shapes.symbols}

\usepackage{MnSymbol}
\definecolor{orange}{rgb}{1.0,0.4,0.0}
\usetikzlibrary{graphs}

\definecolor{MODE}{rgb}{0.98, 0.91, 0.71}

\definecolor{amethyst}{rgb}{0.6, 0.4, 0.8}
\definecolor{asparagus}{rgb}{0.53, 0.66, 0.42}
\definecolor{airforceblue}{rgb}{0.36, 0.54, 0.66}
\definecolor{burlywood}{rgb}{0.87, 0.72, 0.53}
\definecolor{buff}{rgb}{0.94, 0.86, 0.51}
\definecolor{brightlavender}{rgb}{0.75, 0.58, 0.89}
\definecolor{MyBlue}{rgb}{0.25,0.5,0.75}
\definecolor{arylideyellow}{rgb}{0.91, 0.84, 0.42}
\definecolor{cadetgrey}{rgb}{0.57, 0.64, 0.69}
\definecolor{arsenic}{rgb}{0.23, 0.27, 0.29}
\definecolor{tangerine}{rgb}{0.95, 0.52, 0.0}

\usepackage{url}

\usepackage{microtype}

\usepackage{amsmath}

\usepackage{amsmath}

\begin{document}
\title{Self-Supervised Image-to-Text and Text-to-Image Synthesis}
\titlerunning{Self-Supervised Image-to-Text and Text-to-Image Synthesis}
%
\author{Anindya Sundar Das(\Letter) \and
Sriparna Saha}
\authorrunning{A. Das \and S. Saha}
%
\institute{Department of Computer Science and Engineering\\Indian Institute of Technology Patna, India\\
\email{anindyasd34, sriparna.saha@gmail.com}\\}
%
\maketitle              
\setcounter{footnote}{0}
\begin{abstract}

A comprehensive understanding of vision and language and their interrelation are crucial to realize the underlying similarities and differences between these modalities and to learn more generalized, meaningful representations. In recent years, most of the works related to Text-to-Image synthesis and Image-to-Text generation, focused on supervised generative deep architectures to solve the problems, where very little interest was placed on learning the similarities between the embedding spaces across modalities. In this paper, we propose a novel self-supervised deep learning based approach towards learning the cross-modal embedding spaces; for both image to text and text to image generations. In our approach, we first obtain dense vector representations of images using StackGAN-based autoencoder model and also dense vector representations on sentence-level utilizing LSTM based text-autoencoder; then we study the mapping from embedding space of one modality to embedding space of the other modality utilizing GAN and maximum mean discrepancy based generative networks. We, also demonstrate that our model learns to generate textual description from image data as well as images from textual data both qualitatively and quantitatively.

\end{abstract}
\section{Introduction}
The web contains a multitude of images; most of the content images are unannotated. Describing the content of an image automatically using proper natural languages is a vital task, relevant to the area of both Natural Language Processing and Computer Vision, the impact of which could be significant, for example, it will help visually impaired people to have a better understanding of the content of images on the web using existing text-to-speech systems. It has many other important applications such as semantic visual search \cite{fan2011semantic}, or visual intelligence in chatbots \cite{das2017visual}. The reverse problem is the generation of realistic images from human-written descriptions. Although notable progress has been made in generating visually realistic images, those are still far from this goal. The Generative Adversarial Network (GAN) \cite{goodfellow2014generative} based models showed promising results by generating many probable visual representations of a given textual description \cite{reed2016generative}, \cite{reed2016learning}, \cite{zhang2018stackgan++}.

However, one major problem in both text to image synthesis and image to text generation is that recent state-of-the-art deep models are supervised learning-based that require annotated data. Most of the available web data is unlabeled and requires expensive human annotations. Similar problems have been addressed in machine translation where unsupervised machine translation \cite{lample2017unsupervised} attains astounding results even in cases of low-resource languages and also performs reasonably well in the case of distant languages \cite{lample2018phrase}. Motivated by the recent success in machine translation, we aim to investigate whether it is possible to learn the cross-modal embedding spaces between text and visual data in a self-supervised fashion.

\textit{The major contributions of this paper can be summarized as follows:}
    \textit{\textbf{i.} To the best of our knowledge, the current work is the first attempt in developing a self-supervised or unsupervised way of generating images from texts and texts from images.
    \textbf{ii.} Firstly a generative deep image autoencoder setup is developed for generating compelling image vector semantic space as per the reconstructions are concerned. Secondly,  a recurrent neural network based autoencoder is developed which embeds sentences in textual semantic space. Finally, the mapping between cross-modal semantic spaces is established using both GANs and  Maximum Mean Discrepancy (MMD) \cite{tolstikhin2016minimax} based generative networks \cite{li2017mmd} which utilize the adversarial kernel learning technique.
    \textbf{iii.} Results on Caltech-UCSD Birds-200-2011 and Oxford-102 datasets illustrate that self-supervised Image-to-Text and Text-to-Image generation techniques can generate one modality given the other modality.}

\section{Related Works}\label{related}
Caption generation using neural networks was first proposed in \cite{kiros2014multimodal} which used a multimodal log-bi-linear model. In the paper \cite{vinyals2015show}, the authors used deep convolution neural networks (CNN) as an image encoder while RNN as a decoder that generates captions. The authors of \cite{xu2015show}, introduced an attention-based approach for image caption generation which uses convolutional networks for image feature extraction and attention-based RNN as decoder for caption generation. A deep visual semantic captioning model is proposed in \cite{venugopalan2017captioning} that takes the advantage of external sources and exploits the semantic information to generate captions and can describe the objects not present in image-caption datasets. A caption generation model that is based on the dependencies between caption words, image regions, and RNN language model has been proposed in \cite{pedersoli2017areas}. The authors \cite{lao2019dual} showed in their work, that the meaningful style representation which is not well-described in text, but present in the images can be learned in an unsupervised manner. There have been several works on the generation of text for a given image, but there are also the ones that generate an image for a given text. Generative Adversarial Networks(GANs) \cite{goodfellow2014generative} are proven to be useful in generating photo-realistic images \cite{reed2016generative} \cite{zhang2017stackgan}. For text to image task, StackGAN-v2 \cite{zhang2018stackgan++} employs a tree-like structure comprising of multiple generators and discriminators. It generates images at multiple scales at different branches of the tree.

Our work is as closely related to machine translation as it is related to image-caption generation. Recent work in unsupervised machine translation \cite{lample2017unsupervised} uses a shared encoder-decoder architecture and maps the sentences from two different languages into common latent space. A work, related to unsupervised learning \cite{li2017mmd} that employs maximum mean discrepancy (MMD) \cite{tolstikhin2016minimax} based on two-sample tests, minimizes the MMD distance between the two distributions to obtain the mapping. Building on the ideas and advances in these previous related works, we propose a novel architecture for cross-modal generations that utilizes GAN and MMD GANs \cite{li2017mmd} for unsupervised learning.

\section{Methods}\label{method}
Our proposed self-supervised framework comprises of three sub-modules:
1) A deep StackGAN \cite{zhang2018stackgan++} based image autoencoder 2) LSTM-based Sequence-to-Sequence text autoencoder model 3) A cross-modal embedding space mapper that maps embedding from one semantic space to the other. We make our code publicly available\footnote{https://github.com/anindyasdas/SelfSupervisedImageText}.

\subsection{StackGAN based Image Autoencoder}\label{imageauto}
\begin{figure*}[!t]
\centering
\includegraphics[scale=0.47]{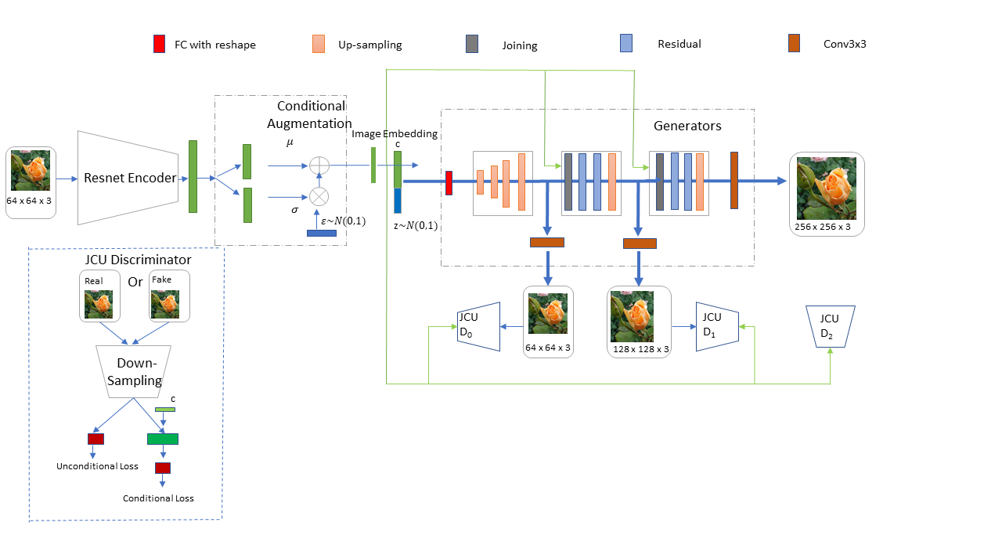}
\caption{Image autoencoder architecture: StackGAN Image Autoencoder. $c$ is conditioning variable} \label{imageae}
\end{figure*}

Our proposed model for image autoencoder is StackGAN based. StackGAN-v2 \cite{zhang2018stackgan++} takes a text embedding vector as input and produces images of increasing resolution at different branches of the network. We modified the architecture to make it an encoder-decoder based network as shown in Figure \ref{imageae}, where an encoder takes an image, extracts different features at different layers of the deep network to obtain an image embedding; this image embedding subsequently is fed as input to the original conditional StackGAN decoder model which reconstructs the image at the output of the conditional StackGAN thus working as an image autoencoder. For the encoder part of the autoencoder, we have used the pre-trained ResNet-50 \cite{he2016deep}. 
We have redefined the last layer of ResNet-50 to obtain $1024$ dimensional image embedding. 

On the decoder side, the generated image is conditioned on the image-embedding vector. As described in Figure \ref{imageae}, firstly the ResNet encoder encodes image $i$ into image embedding $\psi_{i}$. Then Conditional Augmentation technique \cite{zhang2018stackgan++} is applied on image embedding $\psi_{i}$ to produce continuous latent space, yielding condition variable $\hat{c} \sim \mathcal{N}(\mu(\psi_{i}), \sigma(\psi_{i}))$. The following Kullback-Leibler (KL) divergence loss term is optimized during generator training which acts as regularization that ensures the smoothness of the conditioning variable distribution:
\begin{equation}
    D_{KL}(\mathcal{N}(\mu(\psi_{i}), \sigma(\psi_{i})) \|\|\mathcal{N}(0,1))
\end{equation}
 For each generator, $G_{i}$, at different branches of the tree, the hidden feature, $h_{i}$, at $i$-th branch is computed as $h_{0} = Fn_{0}(c, z)$ and
 $h_{i} = Fn_{i}(h_{i-1}, c)$ where $Fn_{i}$ is neural network, $i=0, 1, ..., n-1$; $n$ is the total number of branches. Generators at different stages produce images $u_{i} = G_{i}(h{i})$ from low-to-high resolutions gradually adding more details. Both conditional loss and unconditional loss are being optimized while training the discriminator, $D_i$: 
\begin{multline}
 Loss_{D_{i}}= -\mathbb{E}_{x_{i} \sim p_{data_{i}}}[\log(D_{i}(x_{i})] -\mathbb{E}_{u_{i} \sim p_{G_{i}}}[\log(1 - D_{i}(u_{i})] \\ -\mathbb{E}_{x_{i} \sim p_{data_{i}}}[\log(D_{i}(x_{i}, c)] -\mathbb{E}_{u_{i} \sim p_{G_{i}}}[\log(1 - D_{i}(u_{i}, c)]
\end{multline}
The unconditional loss dictates whether the image at discriminator input is fake or real; the conditional loss decides whether the generated image corresponds to the respective conditioning variable (Figure \ref{imageae}). During the training of the generators, the following loss function is being optimized:
\begin{equation}
 Loss_{G_{i}}= -\mathbb{E}_{u_{i} \sim p_{G_{i}}}[\log(D_{i}(u_{i})] \\  -\mathbb{E}_{u_{i} \sim p_{G_{i}}}[\log(D_{i}(u_{i}, c)]
\end{equation}
The final loss function for training the generators is given by 
\begin{equation}
Loss_{G} = \sum_{0}^{n-1}Loss_{G_{i}}
\end{equation}
\subsection{LSTM-based Sequence-to-Sequence Text Autoencoder Model}\label{textauto} We have used a single-layer LSTM encoder and decoder in our text autoencoder as a sequence-to-sequence model. The encoder is bidirectional LSTM with hidden dimension 50, while the decoder is unidirectional with hidden size 100. The text encoder takes the word embedding vectors at each time step as inputs and generates corresponding hidden vectors. In this paper, we have considered all the hidden vectors to obtain the sentence embedding by max-pooling over all the hidden vectors. This latent vector is used to initialize the decoder LSTM network which regenerates the text description at the output.

\subsection{Cross-modal Embedding Space Mapping Networks}\label{mapper_0}
Cross-modal embedding space mapping networks map from one modality embedding space to the other modality embedding space. The images and texts need not be paired, as both the networks minimize the distance between the two semantic distributions. We have employed two different architectures: One is GAN-based, and the other utilizes MMD-based generative networks.

\subsubsection{GAN-based Cross-modal Embedding Space Mapping Networks\\}\label{mapper_1}
In this architecture, simple GAN models are used, for both image-to-text and text-to-image conversions. The generator translates one modality embedding into the other modality embedding, and the discriminator determines whether two embedding distributions match or not.
\subsubsection{MMD GAN-based Cross-modal Embedding Space Mapping Networks\\}\label{mapper}
Maximum Mean Discrepancy (MMD) is a distance measure on the embedding probability space in Reproducing Kernel Hilbert Space (RKHS) \cite{berlinet2011reproducing}. Given two probability distributions P and Q and a continuous positive definite real-valued kernel $k$ ($\mathcal{H}$ to be corresponding RKHS); the corresponding kernel means be defined as $\mu_{p}= \int k(.,x)dP(x)$ and $\mu_{q}= \int k(.,y)dQ(y)$, the distance $MMD(P,Q) = \|{\mu_{p} - \mu_{q}}\|$ measures the similarity between the two distributions, is known as MMD \cite{gretton2012kernel}. In this paper, we propose a mapping network based on MMD-GAN \cite{li2017mmd} that trains the generator $g_{\theta}$ to minimize MMD distance between two distributions, i.e.\ $\min_{\theta}M_{k}(\mathbb{P}_{\mathcal{X}}, \mathbb{P}_{\mathcal{\theta}}) $, hence passes the hypothesis test.

Here real data $x \sim \mathbb{P}_{\mathcal{X}}$, generator distribution $g_{\theta}(z) \sim \mathbb{P}_{\mathcal{\theta}}$ and $\mathbb{P}_{\mathcal{Z}}$ is the base distribution such that $z \sim \mathbb{P}_{\mathcal{Z}}$ and $M_{k}$ is the square of MMD distances :
\begin{multline}\label{eqn:mmd2}
    M_{k}(\mathbb{P}_{\mathcal{X}}, \mathbb{P}_{\mathcal{\theta}}) =\|\mu_{\mathbb{P}_{\mathcal{X}}} - \mu_{\mathbb{P}_{\mathcal{\theta}}}\|^{2}\\
    =\mathbb{E}_{\mathbb{P}_{\mathcal{X}}}\{k(x, x')\}  +   \mathbb{E}_{\mathbb{P}_{\mathcal{\theta}}}\{k(g_{\theta}(z), g_{\theta}(z'))\} \\
    - 2\mathbb{E}_{\mathbb{P}_{\mathcal{X}}, \mathbb{P}_{\mathcal{\theta}}}\{k(x, g_{\theta}(z))\}
\end{multline}
\begin{figure*}[!t]
    \centering
    \includegraphics[scale=0.42]{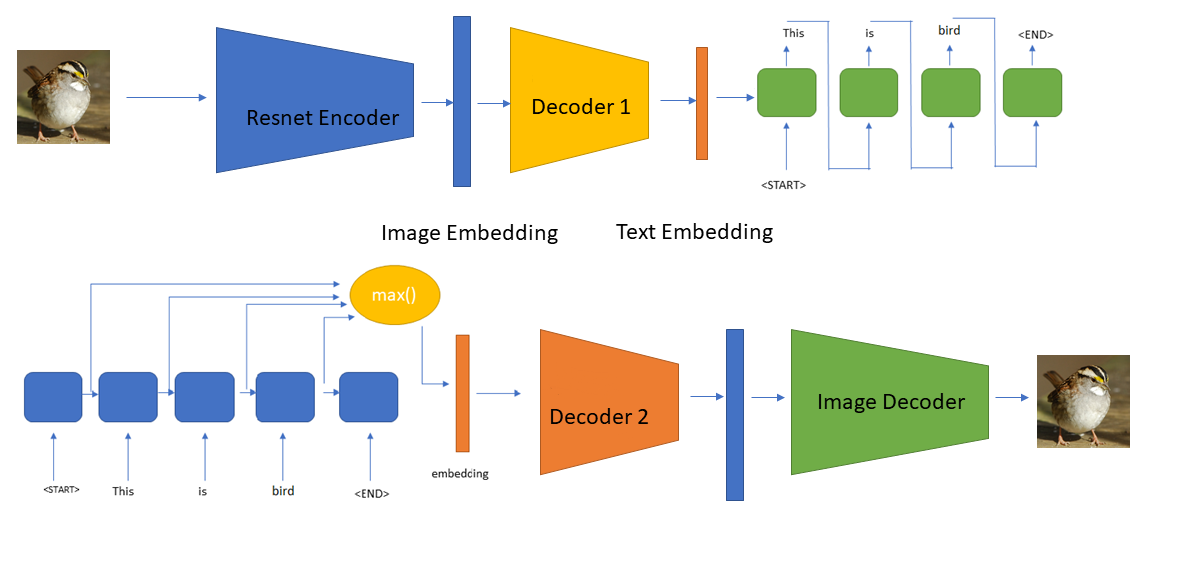}
\caption{End-to-end Networks: Self-supervised Image-to-Text Generator and Self-supervised Text-to-Image Generator}\label{endtoendfig}
\end{figure*}
\section{Experiment}
We have compared our MMD-based Cross-modal mapping network with a GAN-based mapping network, in which the generator generates a mapping from one embedding space to another, and the discriminator is a simple network to distinguish real and fake embeddings. 

We first trained our image autoencoder (Section \ref{imageauto}) for 600 epochs. We used pre-trained ResNet-50 encoder \cite{he2016deep}, the last layer of ResNet is re-defined as a fully connected layer with output dimension 1024, which is also our image vector dimension. During the training of the entire autoencoder, the weight of the ResNet, except the last layer, is kept fixed, while the parameters of the remaining network, are set as trainable. As the ResNet model is pre-trained on Imagenet dataset \cite{deng2009imagenet}, we resized images to $224 \times 224$ and normalized images of our dataset using the mean and standard deviation of the Imagenet dataset.

Next, we pre-train our LSTM-based text autoencoder (Section \ref{textauto}) on 1 million sentences extracted from One Billion Word Benchmark dataset \cite{chelba2013one}. For initialization of our text autoencoder model, we have used top 50k 100-dimensional Glove embeddings \cite{pennington2014glove} along with Glove embeddings of other unique words in caption datasets. The model was pre-trained for 50 epochs; this pre-trained model worked as an initializer; the model was then trained on all the captions of train sets. We have used Cross-Entropy between the generated and target sentences as the loss function.
 
After we trained both our autoencoders, the GAN-based Mapping Networks (Section \ref{mapper_0}) and the MMD-GAN Mapping Networks (Section \ref{mapper}) are trained separately using the unpaired image and text data, for both the image-to-text and the text-to-image conversions (Figure \ref{endtoendfig}). The weights of the trained autoencoders are kept fixed while we train the mapping networks. In image-to-text setup, first, the image encoder encodes an input image into image embedding, which is then used as input to image-to-text embedding mapping networks, that generate corresponding text embedding, the batch of generated text embeddings are used as fake text embeddings while the text embeddings obtained directly from text encoder act as true text embeddings; fake embeddings and true embeddings are then used to train the generative networks. Likewise, the text-to-image setup is trained.
\paragraph*{\textbf{Baselines: }}
We have compared our StackGAN image autoencoder with the following baselines:

\textbf{\textit{ResNet autoencoder}:} In this architecture we fine tuned ResNet-50 as encoder and the decoder comprises of several layers of upsampling followed by deconvolution layer as used in \cite{zhang2017stackgan}. The latent dimension is set to $1024$.

\textbf{\textit{Com-Rec autoencoder}:} This model is based on \cite{jiang2017end} which is completely CNN based. We added linear layers in the bottleneck to obtain dense vector (dimension $1024$) latent space. 

\textbf{\textit{Compressive autoencoder}:} This architecture is based on model as discussed in the paper \cite{theis2017lossy} which comprises of an encoder, a decoder and a probabilistic model. We have added a linear layer in the bottleneck to obtain dense $1024$-dimensional vector embeddings.

For Mapping Networks, as this is a novel self-supervised approach with no existing baselines available for comparisons, we define the following baselines to evaluate the performance:

\textbf{\textit{GAN-based Cross-modal Mapping Network}:} In this architecture simple GAN models are used for both the image-to-text and text-to-image conversions. The generator translates one modality embedding into the other modality embedding and the discriminator determines whether two embedding distributions match or not.

\textbf{\textit{MMD GAN-based Cross-modal Mapping Network}:} This is statistics based network that uses MMD \cite{gretton2012kernel} as objective function (Section \ref{mapper}).
\paragraph*{\textbf{Dataset: }}
For the task, we have used publicly available standard Caltech-UCSD Birds-200-2011 Dataset \cite{WahCUB_200_2011} and Oxford-102 Flower Dataset. The CUB-200-2011 contains a total of 11,788 images of 200 bird species, with multiple annotations per image. All attributes are visual; mostly related to the shape, color, or pattern of the particular part. We split the dataset into train set (8,855 images, 88540 captions, 150 classes) and test set (2,933 images, 29330 captions, 50 classes) such that the respective classes do not overlap. The object-image size ratio in the dataset is less than 0.5 for 80\% of the images, so we pre-processed the images to maintain the size ratio at values greater than 0.75.

The Oxford-102 dataset contains a total of 8,189 images of 102 categories of flowers. The dataset is split into train set (7,034 images, 70340 captions, 82 classes) and test set (1,155 images, 11,550 captions, 20 classes).
\paragraph*{\textbf{Evaluation Metrics: }}\label{eval}
 Inception score (IS) \cite{salimans2016improved} and Fr\'echet Inception distance (FID) \cite{heusel2017gans} to evaluate our generative image autoencoder model quantitatively.

We have used four standard metrics for the performance evaluation of our LSTM-based text autoencoder. These metrics are BLEU \cite{papineni2002bleu}, METEOR \cite{banerjee2005meteor}, ROUGE\-L \cite{lin-2004-rouge}, CIDEr \cite{vedantam2015cider} to quantify the similarity between the ground reference sentence and autoencoder generated sentence. The official MSCOCO caption evaluation scripts\footnote{https://github.com/tylin/coco-caption
} are used for evaluation.

The following evaluation metrics are used to assess the performance of our mapping networks:
\begin{enumerate}
\item \textbf{\textit{Human Score:}} We conducted user studies for evaluation of our end-to-end text-to-image and image-to-text systems. We sampled 2000 samples for each of the text-to-image and image-to-text cases, assigned 5 different users (excluding any of the authors) to score the results. Generated samples are evaluated on a $4.0$ scale, $1.0$ being the lowest (worst case) and $4.0$ being the highest (best case). Point $4.0$ is awarded when the text description matches completely the image without any errors.  Point $3.0$ is awarded when the text description partially matches the image with minor errors. Point $2.0$ is awarded when there is somewhat matching between text and image. Point $1.0$ is awarded when the image and text are unrelated.

\item \label{subsubsec:1} \textbf{\textit{Class Accuracy:}}
    The CUB-200-2011 dataset has $150$ train classes and $50$ test classes and the Oxford-102 dataset has $82$ and $20$ train and test classes, respectively. Classes define different species for birds. Birds in the same class are from the same species, have identical features and descriptions. It is logical to expect that when one modality embedding maps into the other modality embedding, they are mapped to the same class. We first obtain the embeddings of one modality (say modality \textbf{A}) call it \textbf{true embeddings in modality A}. Then we obtain the embeddings of other modality (say modality \textbf{B}) and use them to feed into the \textbf{"Modality B-To-Modality A" mapping network} to obtain the \textbf{fake embeddings in modality A}. Then the cosine similarity between the true embeddings and fake embeddings are computed and we do an $argmax$ to determine the class of fake embedding based on the highest cosine similarity scores. Then we calculate the class accuracy based on class labels of the true embeddings and predicted class labels of the fake embeddings.
    
\end{enumerate}

\section{Result and Analysis}
The Inception score and FID of the StackGAN autoencoder model have been reported and compared with the baselines in Table \ref{tab:score1}. All the reported results are statistically significant. Results clearly illustrate the best performance by StackGAN autoencoder.




\begin{table*}[h]
\begin{center}
\begin{tabular}{ | p{2.7cm} | p{4.2cm}| p{3.2cm}| p{1.5cm}| }
\hline
\textbf{Dataset} & \textbf{Model} & \textbf{Inception Score} $\uparrow$ & \textbf{FID} $\downarrow$\\
\hline
CUB & \textbf{ResNet autoencoder with latent dimension $1024$} &  1.08 $\pm$ 0.03 & 275.71\\ 
\hline
CUB & \textbf{Com-Rec autoencoder with latent dimension $1024$} & 1.15 $\pm$ 0.02  & 248.95\\
\hline
CUB & \textbf{Compressive autoencoder with latent dimension $1024$} & 1.09 $\pm$ 0.08  & 283.64\\
\hline
CUB & \textbf{StackGAN autoencoder} & \textbf{3.66 $\pm$ 0.09} & \textbf{21.58}\\
\hline
Oxford-102 & \textbf{StackGAN autoencoder} & \textbf{3.23 $\pm$ 0.14} & \textbf{57.96}\\
\hline
\end{tabular}
\end{center}
\caption{Inception scores, Fr\'echet Inception distance (FID) (computed for 256$\times$256 images)}\label{tab:score1}
\end{table*}

 The scores for LSTM-based text autoencoder model are reported as: on CUB dataset (BLEU-1 :87.52, BLEU-4: 72.55, METEOR: 49.89, ROUGE-L : 89.04, CIDEr: 7.04) and on Oxford-102 (BLEU-1 :85.35, BLEU-4: 70.61, METEOR: 48.06, ROUGE-L : 86.69, CIDEr: 6.59)

\begin{figure*}[!h]
    \centering
    \includegraphics[scale=0.35]{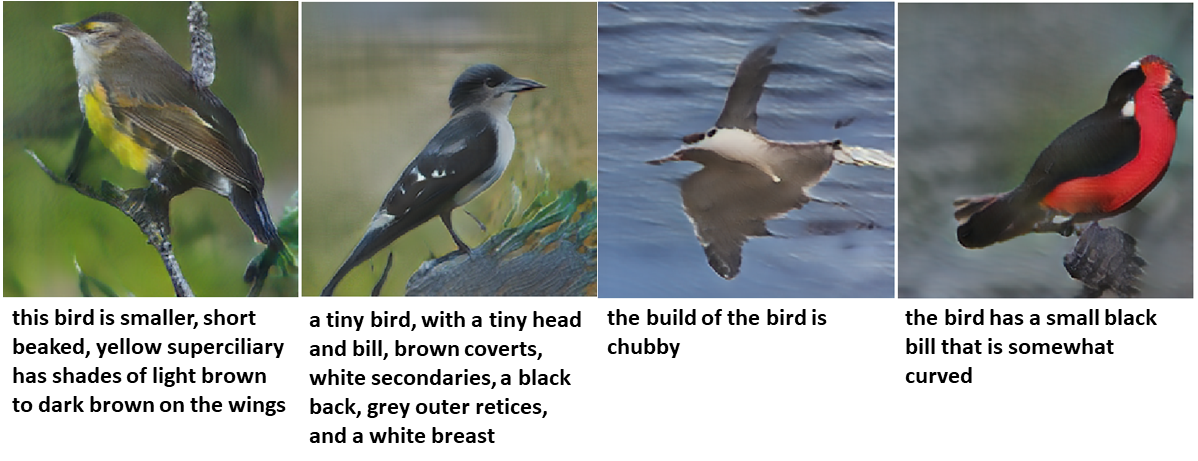}
\caption{End-To-End Network Output for Birds: Text-To-Image Synthesis}\label{gan_txt_img_bird}
\end{figure*}

\begin{figure*}[!t]
    \centering
    \includegraphics[scale=0.35]{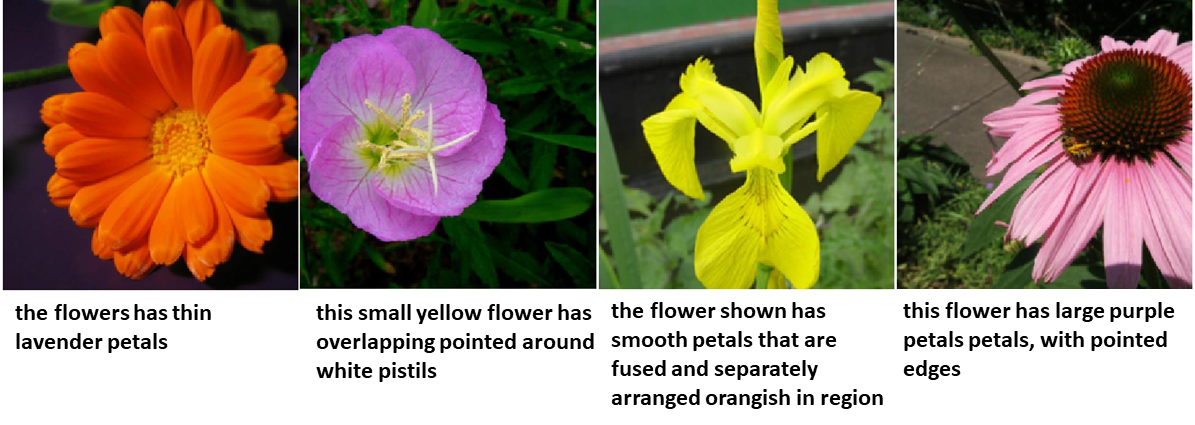}
\caption{End-To-End Network Output for Flowers: Image-To-text Synthesis}\label{gan_img_txt_flower}
\end{figure*}

The sample outputs of the End-To-End Image-To-Text synthesis are shown in Figure  \ref{gan_img_txt_flower}; Figure \ref{gan_txt_img_bird} depicts the sample outputs for Text-to-Image synthesis network. Evaluation metric scores are reported in Table \ref{tab:endresults_gan} for GAN-based end-to-end networks and MMD-based networks.

The human score indicates that there exist some correlations between the two modalities of Image-to-Text and Text-to-Image synthesis systems. However, \textbf{Class Accuracy} score is indicative of a low semantic correlation between the two modalities for both GAN based and MMD based Image-to-Text and Text-to-Image systems. The scores at Table \ref{tab:endresults_gan} imply similar performances by both the GAN-based system and the MMD-based system.




\begin{table*}[!h]
\begin{center}
\begin{tabular}{| p{1.9cm}| p{1.7cm} | p{2cm} | p{2cm} | p{2cm} | p{2cm} |}
\hline
\multirow{2}{*}{Model} &
\multirow{2}{*}{Metric} &  \multicolumn{2}{c|}{CUB-200-2011} & %
    \multicolumn{2}{c|}{Oxford-102} \\
\cline{3-6}
  & & Image-To-Text & Text-To-Image & Image-To-Text & Text-To-Image \\
\hline
\multirow{2}{*}{GAN-based} 
& Human Score & \textbf{2.3} & \textbf{2.5} & \textbf{2.5} & \textbf{2.4} \\
\cline{2-6}
 & Class Accuracy* & \textbf{2.3}  & \textbf{2.8} & \textbf{6.6} & \textbf{4.5} \\
\hline
 
\multirow{2}{*}{MMD-based} &
Human Score & 2.1  & 2.4 & 2.2 & 2.3 \\
\cline{2-6}
&  Class Accuracy* & 2.0 & 2.5 & 5.6 & 4.1 \\
\hline

\end{tabular}
\end{center}
\caption{Human Score, Class Accuracy of GAN-based and MMD-based Image-To-Text and Text-To-Image End-To-End systems on CUB and Oxford-102 datasets. (* means the values reported are in percentage).}\label{tab:endresults_gan}
\end{table*}



\section{Error Analysis}
A thorough analysis revealed scenarios for possible reasons for errors. The strength of the end-to-end model is dependant on strength of the individual components, as models are trained separately; so error in each component has a cumulative impact on the end-to-end network. The autoencoders are trained separately without sharing weights or information, as a result, the latent spaces are disjoint, which further makes it harder for the mapping networks to align learned embedding spaces on the semantic level.
\section{Conclusion}
In this paper, we proposed a novel self-supervised approach that performs the cross-modal translation from image to text and text to image. We showed that our Text-to-Image and Image-to-Text synthesis networks learn to map the semantic space of one modality to the semantic space of the other modality in an unsupervised fashion. However, we figure out while learning the mapping, the semantic correlation across the modalities is low. Though the current end-to-end network depicts low cross-modal semantic alignments, as part of the future work, these learned network weights can be used as initialization for the synthesis networks and the entire network can be fine-tuned till the embeddings of the two modalities semantically align in the latent space. 

\section*{Acknowledgments}
Dr. Sriparna Saha gratefully acknowledges the Young Faculty Research Fellowship (YFRF) Award, supported by Visvesvaraya Ph.D. Scheme for Electronics and IT, Ministry of Electronics and Information Technology (MeitY), Government of India, being implemented by Digital India Corporation (formerly Media Lab Asia) for carrying out this research.


\end{document}